\documentclass[10pt,twocolumn,letterpaper]{article}

\usepackage{cvpr}
\usepackage{times}
\usepackage{epsfig}
\usepackage{graphicx}
\usepackage{amsmath}
\usepackage{amssymb}

\usepackage{enumerate}
\usepackage{adjustbox}
\usepackage{multirow}
\usepackage{multicol}
\usepackage{booktabs}
\usepackage[dvipsnames]{xcolor}

\usepackage{pifont}
\usepackage{mathtools}
\usepackage{extarrows}
\usepackage{comment}
\usepackage{makecell}
\usepackage{float}  
\usepackage{multirow}
\usepackage{fixmath}
\usepackage{amsfonts}
\usepackage[nocompress]{cite}  

\usepackage{xspace}

\makeatletter
\DeclareRobustCommand\onedot{\futurelet\@let@token\@onedot}
\def\@onedot{\ifx\@let@token.\else.\null\fi\xspace}
\def\eg{\emph{e.g}\onedot} 
\def\ie{\emph{i.e}\onedot}

\def\wrt{w.r.t\onedot} 
\def\etal{\emph{et al}\onedot}

\makeatother

\newcommand{\partitle}[1]{\noindent\textbf{#1}}

\newcommand{\B}[1]{\textbf{#1}}

\newcommand{\vt}[1]{\boldsymbol{#1}}

\usepackage[breaklinks=true,bookmarks=false]{hyperref}

\cvprfinalcopy 

\def\cvprPaperID{4686} 

\ifcvprfinal\fi
\begin{document}

\title{Spherical Regression:\\ Learning Viewpoints, Surface Normals and 3D Rotations on \emph{n}-Spheres}

\author{
    Shuai Liao \\
    \and
    Efstratios Gavves\\
    %
    \and
    Cees G. M. Snoek\\
    \and
    \and
    QUVA Lab, University of Amsterdam
}


\maketitle
\thispagestyle{empty}

\begin{abstract}
Many computer vision challenges require continuous outputs, but tend to be solved by discrete classification. The reason is classification's natural containment within a probability $n$-simplex, as defined by the popular softmax activation function. Regular regression lacks such a closed geometry, leading to unstable training and convergence to suboptimal local minima. Starting from this insight we revisit regression in convolutional neural networks. We observe many continuous output problems in computer vision are naturally contained in closed geometrical manifolds, like the Euler angles in viewpoint estimation or the normals in surface normal estimation. A natural framework for posing such continuous output problems are $n$-spheres, which are naturally closed geometric manifolds defined in the $\mathbb{R}^{(n+1)}$ space. By introducing a spherical exponential mapping on $n$-spheres at the regression output, we obtain well-behaved gradients, leading to stable training. We show how our spherical regression can be utilized for several computer vision challenges, specifically viewpoint estimation, surface normal estimation and 3D rotation estimation. For all these problems our experiments demonstrate the benefit of spherical regression. All paper resources are available at {\color{gray}\href{https://github.com/leoshine/Spherical_Regression}{https://github.com/leoshine/Spherical\_Regression}}.
\end{abstract}

 %


\section{Introduction}          \label{sec:intro}

Computer vision challenges requiring continuous outputs are abundant.
Viewpoint estimation~\cite{tulsiani2015viewpoints,su2015render,penedones2012improving, prokudin2018deep},
object tracking~\cite{tao2016sint,gao2014transfer,hong2015online,kalal2012tracking}, and surface normal estimation~\cite{bansal2016marr,eigen2015predicting,zhang2017physically,qi2018geonet} are just three examples. Despite the continuous nature of these problems, regression based solutions that seem a natural fit are not very popular. Instead, classification based approaches are more reliable in practice and, thus, dominate the literature ~\cite{tulsiani2015viewpoints,su2015render,massa2016crafting, tao2016sint, LadickySP14}.
This leads us to an interesting paradox: while several challenges are of continuous nature, their present-day solutions tend to be discrete.

\begin{figure}[t!]
\centering
    \includegraphics[width=1.0\linewidth]{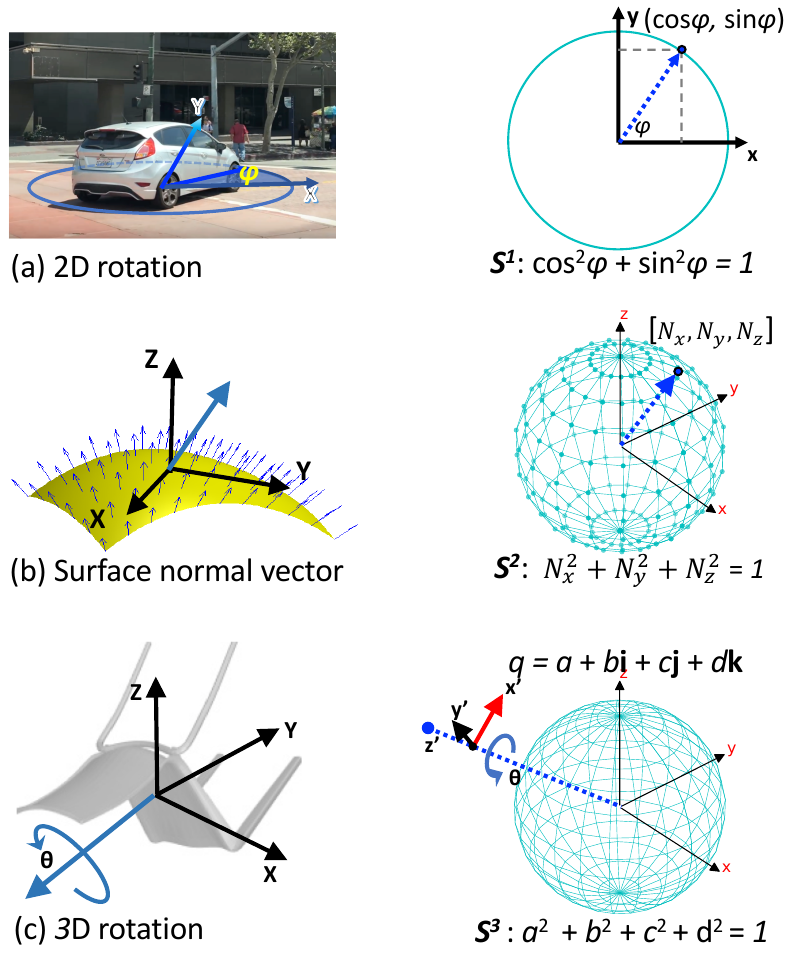}
   \caption{\textbf{Many computer vision problems can be converted into a $n$-sphere problem.}
    $n$-spheres are naturally closed geometric manifolds defined in the $\mathbb{R}^{(n+1)}$ space.
    Examples are a) viewpoint estimation, b) surface normal estimation, and c) 3D rotation estimation.
    This paper proposes a general regression framework that can be applied on all these $n$-sphere problems.
   }
\label{fig:use_case}
\end{figure}

In this work we start from this paradox and investigate why regression lags behind. When juxtaposing the mechanics of classification and regression we observe that classification is naturally contained within a probability $n$-simplex geometry defined by the popular softmax activation function.
The gradients propagated backwards to the model are constrained and enable stable training and convergence.
In contrast, regression is not contained by any closed geometry. Hence, the gradients propagated backwards are not constrained, potentially leading to unstable training or convergence to suboptimal local minima.
Although classification solutions for continuous problems suffer from discretization errors in annotations and predictions, they typically lead to more reliable learning ~\cite{massa2016crafting, LadickySP14}.

Founded on the relation between classification, regression and closed geometric manifolds, we revisit regression in deep networks. Specifically, we observe many continuous output problems in computer vision are naturally contained in closed geometrical manifolds defined by the problem at hand. For instance, in viewpoint estimation, angles cannot go beyond the $[-\pi, \pi]$ range. Or, in surface normal estimation the $\ell_2$ norm of the surface normals must sum up to 1 to form unit vectors that indicate directionality. It turns out that a natural framework for posing such continuous output problems are the $n$-spheres $S^n$ ~\cite{flanders1963differential,eduarda1996experiencing}, which are naturally closed geometric manifolds defined in the $\mathbb{R}^{(n+1)}$ space. We, therefore, rethink regression in continuous spaces in the context of $n$-spheres, when permitted by the application. It turns out that if we introduce a proposed spherical exponential mapping on $n$-spheres  at the regression output we obtain regression gradients that are constrained and well-behaving, similar to classification-based learning. We refer to regression using the proposed spherical exponential mappings on $S^n$ spheres as $S^n$ spherical regression.

In this work we make the following contributions. First, we link the framework of $n$-spheres to continuous output computer vision tasks. By doing so, they are amenable to the properties of the $n$-spheres formulation, leading to spherical regression. Second, we propose a novel nonlinearity, the spherical exponential activation function, specifically designed for regressing on $S^n$ spheres. We show the activation function improves the results obtained by regular regression. Third, we show how the general spherical regression framework can be utilized for particular computer vision challenges. Specifically, we show how to recast existing methods for viewpoint estimation, surface normal estimation and 3D rotation estimation to the proposed spherical regression framework. Our experiments demonstrate the benefit of spherical regression for these problems.

We now first describe in Section ~\ref{sec:problem} the motivation behind the deep learning mechanics of classification and regression.
Based on the insights derived, we describe in Section ~\ref{sec:model} the general framework for spherical regression on $S^n$ spheres.
We then explain how to specialize the general frameworks for particular applications, see Fig.~\ref{fig:use_case}. We describe the related work for these tasks in Section ~\ref{sec:related}. In Section ~\ref{sec:experiments}, we evaluate spherical regression for the three applications.

\section{Motivation}            \label{sec:problem}
\partitle{Deep classification and regression networks.} We start from an input image $\mathbf{x}$ of an object with a supervised learning task in mind, be it classification or regression. Regardless the task, if we use a convolutional neural network (CNN) we can split it into two subnetworks, the base network and the prediction head, see (eq. \ref{eq:network}).

\begin{equation}
\underbrace{
    \mathbf{x} \xrightarrow[\text{base network}]{H(\cdot)}
    \vt O =  \left[
        \begin{array}{ccc}
        o_0    \\
        o_1    \\
        \vdots \\
        o_n    \\
        \end{array}
        \right]
}_{\text{Base network}}
\xrightarrow[\text{activation}]{g(\cdot)}
\underbrace{
    \vt P =  \left[
        \begin{array}{ccc}
        p_0    \\
        p_1    \\
        \vdots \\
        p_n    \\
        \end{array}
        \right]
    \xleftrightarrow[\text{loss}]{\mathcal{L}(\cdot,\cdot)}
    \vt Y
}_{\text{Prediction head}}
\label{eq:network}
\end{equation}

The base network considers all the layers from input $\mathbf{x}$ till layer $\vt O$. It defines a function $\vt O=H(\mathbf{x})$ that returns an intermediate latent embedding $\vt O=[o_0, o_1, ..., o_n]^\top$ of the raw input $\mathbf{x}$. The function comprises a cascade of convolutional layers intertwined with nonlinearities and followed by fully connected layers, $H=h_l \circ h_{l-1} \dots \circ h_k \circ \dots \circ h_2 \circ h_1$, where $h_k$ is the $\theta$-parameterized mapping of $k$-th layer. Given an arbitrary input signal $\mathbf{x}$, the latent representation $\vt O$ is unconstrained, namely $\mathbf{x} = H(\mathbf{x}) \rightarrow \mathbb{R}^{(n+1)}$.

The prediction head contains the last ($n+1$)-dimensional layer $\vt P$ before the loss function, which is typically referred to as the network output. The output is obtained from an activation function $g(\cdot)$, which generates the output $\vt P: p_k=g(o_k; \vt O)$ using as input the intermediate raw embedding $\vt O$ returned by the base network. The activation function $g(\cdot)$ imposes a structure to the raw embdedding $\vt O$ according to the task at hand.
For instance, for a CNN trained for image classification out of $1,000$ classes we have a $1,000$-dimensional output layer $\vt P$ that represents softmax probabilities.
And, for a CNN trained for 2D viewpoint estimation we have a $2$-dimensional output layer $\vt P$ that represents the trigonometric functions $\vt P=[cos\phi,sin\phi]$.
After the prediction head lies the loss function ${\mathcal{L}(\vt P, \vt Y)}$ that computes the distance between the prediction $\vt P$ and the ground truth $\vt Y=[y_0, y_1, ...]^\top$, be it cross entropy for classification or sum of squared errors for regression.

The dimensionalities of $\vt O$ and $\vt P$ vary according to the type of classification or regression that is considered. For classification $\vt P$ represents the probability of ($n+1$) discretized bins. For regression, $\vt P$ depends on the assumed output representation dimensionality, \eg, regression 1D ~\cite{penedones2012improving}, regression 2D ~\cite{penedones2012improving, beyer2015biternion} or regression 3D ~\cite{osadchy2007synergistic} and beyond can have different output dimensions. Together the subnetworks comprise a standard deep architecture, which is trained end-to-end.

\partitle{Training.}
During training, the $k$-th layer parameters are updated with stochastic gradient descent,
$\theta_k \leftarrow \theta_k - \gamma \frac{\partial \mathcal{L}}{\partial \theta_k}$,
where $\gamma$ is the learning rate. Expanding by the chain rule of calculus we have that
\begin{equation}
\frac{\partial \mathcal{L}}{\partial \theta_k}
= \frac{\partial \mathcal{L}}{\partial \vt P}
\frac{\partial \vt P}{\partial \vt O}
    (\frac{\partial \vt O}{\partial h_{l-1}} \dots
    \frac{\partial h_{k+1}}{\partial h_k})
    \frac{\partial h_k}{\partial \theta_k}
    \label{eq:chain}
\end{equation}
Training is stable and leads to consistent convergence when the gradients are constrained,
otherwise gradient updates may cause bouncing effects on the optimization landscape and may cancel each other out.
Next, we examine the behavior of the output activation $\vt P$ and the loss functions for classification and regression.

\partitle{Classification.}
For classification the standard output activation and loss functions are the softmax and the cross entropy,
that is $g(o_i; \vt O) = \{ p_i = e^{o_i}/\sum_j e^{o_j}, i=0 \cdots n \}, \mathcal{L}(\vt O,\vt Y) = - \sum_i y_i log(p_i)$. The $p_i$ and $y_i$ are the posterior probability and the one-hot vector for the $i$-th class, and $d$ is the number of classes.
Note that softmax maps the raw latent embedding $O \in \mathbb{R}^{(n+1)}$ to a structured output $\vt P$, known as $n$-simplex,
where each dimension is positive and the sum equals to one, \ie $\sum_i p_i = 1$ and $p_i>0$.
The partial derivative of the probability output with respect to the latent activation equals to

\begin{equation}
\frac{\partial p_j}{\partial o_i} =
\begin{cases}
     p_j \cdot (1-p_j) , & \text{when $j=i$}      \\
    -p_i \cdot p_j     , & \text{when $j\neq i$}
\end{cases}
\label{eq:p-o-cls}
\end{equation}
%
Crucially, we observe that the partial derivative $\frac{\partial p_j}{\partial o_i}$  does not directly depend on $\vt O$. This leads the partial derivative of the loss function with respect to $o_i$, namely
\begin{equation}
\frac{\partial \mathcal{L}}{\partial o_i}
 = - \sum_k  \frac{y_k}{p_k} \cdot \frac{\partial p_k}{\partial o_i}
 = p_i - y_i,
\end{equation}
to be independent of $\vt O$ itself. As $\vt P$ corresponds to a probability distribution that lies inside the $n$-dimensional simplex, it is naturally constrained by its $\ell_1$ norm, $p_j<1$. Thus, the partial derivative $\frac{\partial \mathcal{L}}{\partial \vt O}$ depends only on a quantity that is already constrained.

\bigbreak\noindent\textbf{Regression.}
In regression usually there is no explicit activation function in the final layer to enforce some manifold structure. Instead, the raw latent embedding $\vt O$ is directly compared with the ground truth. Take the smooth-L1 loss as an example,
\begin{equation}
\mathcal{L}=
\begin{cases}
   0.5 |y_i-o_i|^2 & \text{if} |y_i-o_i| \leqslant 1 \\
   |y_i-o_i| - 0.5 & \text{otherwise}.
\end{cases}
\end{equation}
The partial derivative of the loss with respect to $o_i$ equals to
\begin{equation}
\frac{\partial \mathcal{L}}{\partial o_i}=
\begin{cases}
   - (y_i-o_i)     & \text{if} |y_i-o_i| \leqslant 1 \\
   - \text{sign}(y_i-o_i) & \text{otherwise}.
\end{cases}
\end{equation}
Unlike classification, where the partial derivatives are constrained, for regression we observe that the $\frac{\partial \mathcal{L}}{\partial o_i}$ directly depends on the raw output $\vt O$. Hence, if $\vt O$ has high variance, the unconstrained gradient will have a high variance as well. Because of the unconstrained gradients training may be unstable.

\partitle{Conclusion.}
Classification with neural networks leads to stable training and convergence. The reason is that the partial derivatives $\frac{\partial \mathcal{L}}{\partial \vt P} \cdot \frac{\partial \vt P}{\partial \vt O}$ is constrained, and, therefore, the gradient updates $\frac{\partial \mathcal{L}}{\partial \theta_k}$, are constrained. The gradients are constrained because the output $\vt P$ itself is constrained by the $\ell_1$ norm of the $n$-simplex, $\sum_i p_i = 1$. Regression with neural networks may have instabilities and sub-optimal results during training because gradient updates are unconstrained. We examine next how we can define a similar closed geometrical manifold also for regression. Specifically, we focus on regression problems where the target label $Y$ lives in a constrained $n$-sphere manifold.

\section{Spherical regression}  \label{sec:model}

The $n$-sphere, denoted with $S^n$, is the surface boundary of an $(n+1)$-dimensional ball in the Euclidean space.
Mathematically, the $n$-sphere is defined as $S^{n}=\left\{ \mathbf{x} \in \mathbb {R} ^{n+1}:\left\|\mathbf{x}\right\|=r\right\}$ and is constrained by the $\ell_2$ norm, namely $\sum_i x_i^2 = 1$.
Fig. \ref{fig:use_case} gives examples of simple $n$-spheres, where $S^1$ is the circle and $S^2$ the surface of a 3D ball. Where the $n$-simplex constrains classification by the $\ell_1$ simplex norm, we next present how to constrain regression by the $\ell_2$ norm of an $n$-sphere.

\subsection{Constraining regression with $n$-spheres}  \label{ssub:spherical_regrression}


To encourage stability in training regression neural networks on $S^n$ spheres, one reasonable objective is to ensure the gradients are constrained. To constrain the gradient $\frac{\partial \mathcal{L}}{\partial \vt O}$, we propose to insert an additional activation function in regression after the raw embedding layer $\vt O$.
The activation function should have the following properties.
\begin{enumerate}[I]
  \item {The \emph{output} of the activation, $\vt P=\{p_k\}$, must live on $n$-sphere, namely its $\ell_2$ norm $\sum_{k=1} p_k^2=1$ must be constant, \eg, $cos^2\phi+sin^2\phi=1$.
  This is necessary for spherical targets.}\label{prop:I}

  \item {Similar to classification, the \emph{gradient} $\frac{\partial \mathcal{L}}{\partial \vt O}$ must not directly depend on the input signal.
  That is, $\frac{\partial \mathcal{L}}{\partial \vt O}$ must not depend directly on the raw latent embedding $\vt O \in \mathbb{R}^{(n+1)}$.}\label{prop:II}

\end{enumerate}

To satisfy property \ref{prop:I}, we pick our activation function such that it produces normalized values.
We opt for the $\ell_2$ normalization form: $p_j =g(o_j; \vt O)= \frac{f(o_j)}{\sqrt{\sum_k f(o_k)^2}}$, where $f(\cdot)$ corresponds to any univariate mapping.
The partial derivative of the output with respect to the latent $O$ then becomes:
\begin{align} \label{eq:dev_p_o}
    \frac{\partial{p_j}}{\partial o_i}
    &= \frac{\partial{\left[ \frac{f(o_j)}{ \sqrt{\sum_k f(o_k)^2} } \right] }}{\partial o_i}                 \nonumber\\
    &= \begin{cases*}
          \left(\frac{d f(o_i)}{d o_i} \cdot \frac{1}{A}\right) \cdot (1- {p_i}^2),           &\text{when $j=i$}     \\
          \left(\frac{d f(o_i)}{d o_i} \cdot \frac{1}{A}\right) \cdot (-p_i\cdot p_j),        &\text{when $j\neq i$} \\
       \end{cases*}
\end{align}
where $A=\sqrt{\sum_k f(o_k)^2}$ is the normalization factor.

Still, $\frac{\partial p_j}{\partial o_i}$ is potentially depending on the raw latent embedding $\vt O$ through the partial function derivatives $\frac{d f(o_j)}{d o_i}$ and the normalization factor $A$.
To satisfy property \ref{prop:II} and make $\frac{\partial p_i}{\partial o_j}$ independent from the raw output $\vt O$, and thus constrained, we must make sure that $\left(\frac{d{f(o_j)}}{d o_j} \cdot \frac{1}{A}\right)$ becomes independent of $\vt O$. In practice, there are a limited number of choices for $f(\cdot)$ to satisfy this constraint. Inspired by the softmax activation function, we resort to the exponential map $f(o_i)=e^{o_i}$, where $\frac{d f(o_i)}{d o_i}=f(o_i)$ and $\frac{\partial{f(o_i)}}{\partial o_i} \cdot \frac{1}{A} = \frac{f(o_i)}{A} = p_i$. Thus Eq.~\ref{eq:dev_p_o} is simplified as
\begin{align} \label{eq:dp_do_sph}
    \frac{\partial{p_j}}{\partial o_i} &=
    \begin{cases*}
        p_i \cdot (1- p_i^2),   &\text{when $j=i$}     \\
      - p_i^2 \cdot p_j,        &\text{when $j\neq i$} \\
    \end{cases*}
\end{align}
removing all dependency on $\vt O$.

Since our activation function has a similar form as softmax, which is also known as normalized exponential function, we refer to our activation function as \emph{Spherical Exponential Function}. It maps inputs from $\mathbb{R}^{n+1}$ to the positive domain of the $n$-Sphere,
\ie $S_{exp}(\cdot) : \mathbb{R}^{n+1} \rightarrow \mathbb{S}^{n}_+$:
%
\begin{equation} \label{eq:sph_exp}
  p_j = S_{exp}(o_j;\vt O) = \frac{e^{o_j}}{\sqrt{\sum_k (e^{o_k})^2}}
\end{equation}
Converting Eq. \ref{eq:dp_do_sph} into matrix provides Jacobian as $\mathbf{J}_{S_{exp}} = (\vt{I}-\vt{P}\otimes \vt{P}) \cdot diag(\vt{P})$  where $\otimes$ denotes outer product (see supplementary material for details).
Notice that if we only do $\ell_2$ normalization without exponential, the Jacobian is given as $\mathbf{J}_{\mathcal{S}_{flat}} = (\vt{I}-\vt{P}\otimes \vt{P}) \cdot \frac{1}{||\vt O||}$, which is influenced by the magnitude of $\vt O$ in gradient, which is unconstrained.

\begin{figure}[t!]
\centering
\includegraphics[width=\linewidth]{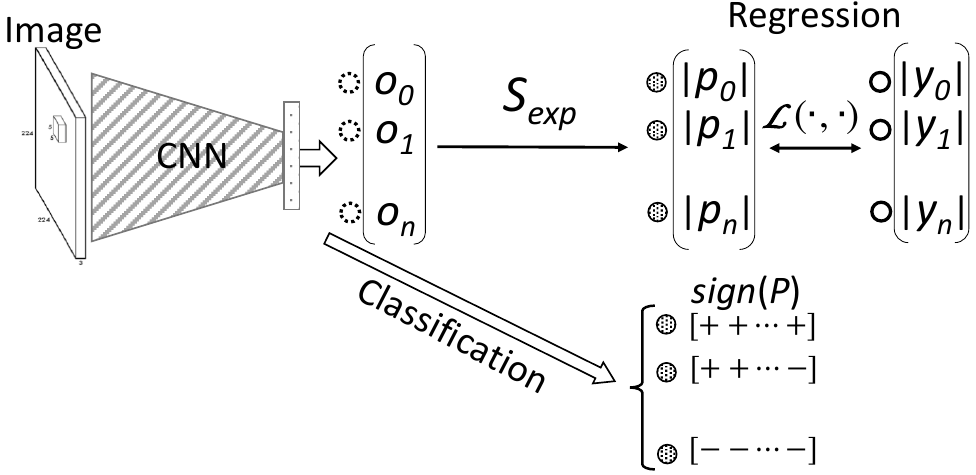}
   \caption{\textbf{Regressing on $n$-spheres} with targets $Y=[y_0, ..., y_n]$, \ie $\sum_i y_i^2=1$.
             The model processes the input image and first returns a raw latent embedding $\vt O =[o_0, ...,o_n] \in \mathbb{R}^{(n+1)}$.
             Then, a regression branch using the proposed spherical exponential acctivation $S_{exp}$
             maps $\vt O$ to a structured output $|\vt P|=[|p_0|,...,|p_n|]$.
             A classification branch is also used to learn the sign labels of $\vt P$.
             Prediction is made by $\vt P=sign(\vt P) \cdot |\vt P|$.}
\label{fig:overall}
\end{figure}

Unfortunately, the exponential map in $S_{exp}(\cdot)$ restricts the output to be in the positive range only, 
whereas our target can be either positive or negative.
To enable regression on the full range on $n$-sphere coordinates we rewrite each dimension into two parts: $p_i=sign(p_i) \cdot |p_i|$.
We then use the output from the spherical exponential function to learn the absolute values $|p_i|, i=0,1,...,n$ only. At the same time, we rely on a separate classification branch to predict the sign values, $sign(p_i), i=1,...,d$ of the output. The overall network is shown in Fig. ~\ref{fig:overall}:
{}

\partitle{Conclusion}.
Given the spherical exponential mapping for $g(\cdot)$,
the gradient $\frac{\partial \vt P}{\partial \vt O}$ is detached from $\vt O$, and $\vt P$ is constrained by the $n$-Sphere.
Thus, to make the parameter gradients also constrained, we just need to pick a suitable loss function.
It turns out that there are no significant constraints for the loss function.
Given ground truth $\vt Y$, we can set the loss to be the negative dot product $\mathcal{L}=-\langle|\vt P|,|\vt Y|\rangle$.
Since both $\vt P$ and $\vt Y$ are on sphere with $\ell_2$ norm equal to 1 (\ie $||\vt P||_2=||\vt Y||_2=1$),
this is equivalent to optimize with
cosine proximity loss or L2 loss \footnote{For cosine proximity loss: $\mathcal{L}=-\frac{\langle|\vt P|,|\vt Y|\rangle}{||\vt P||_2 \cdot ||\vt Y||_2}=-\langle|\vt P|,|\vt Y|\rangle$. For L2 loss: $\mathcal{L}=||\vt P - \vt Y||_2^2 = {||\vt P||}_2^2+{||\vt Y||}_2^2-2\langle|\vt P|,|\vt Y|\rangle = 2 - 2\langle|\vt P|,|\vt Y|\rangle$.}.
In this case, the gradients are $\frac{\partial \mathcal{L}}{\partial{p_i}} = -sign(p_i)|y_i|$ and only relate to $\vt P$.
We could also treat the individual outputs $\{p_1^2, p_2^2\, ...\}$ as probabilities with a cross-entropy loss on continuous labels $y_i^2$, in which case we would have that $H(\vt{Y}^2,\vt{P}^2)=\sum_i y_i^2 \log \frac{1}{p_i^2}$.
We conclude that the Spherical Regression using the spherical exponential mapping allows for constrained parameter updates and, thus, we expect it to lead to stable training and convergence.
We verify this experimentally on three different applications and datasets.

\subsection{Specializing to $S^1, S^2$ and $S^3$}
Next, we show how to specialize the general $n$-sphere formulation for different regression applications that reside on specific $n$-spheres.

\partitle{$S^1$ case: Euler angles estimation.}
Euler angles are used to describe the orientation of a rigid body with respect to a fixed coordinate system.
They are defined by 3 angles, describing 3 consecutive rotations around fixed axes. 
Specifically, each of the angles $\phi \in [0,2\pi]$ can be represented by a point on a unit circle with 2D coordinate $[cos\phi,sin\phi]$, see Fig. \ref{fig:use_case}.
Since $cos^2\phi + sin^2\phi = 1$, estimating these coordinates is an $S^1$ sphere problem.
Consequently, our prediction head has two components: \emph{i)} a regression branch with spherical exponential activations for absolute values $|\vt P| =[ |cos\phi|,|sin\phi| ]$ and, \emph{ii)} a classification branch to learn all possible sign combinations between $sign(cos\phi)$ and $sign(sin\phi)$, that is a 4-class classification problem: $sign(\vt P) \in \{(+,+), (+,-), (-,+), (-,-)\}$. We could also predict the signs independently and have fewer possible outputs, however, this would deprive the classifier from the opportunity to learn possible correlations.

During training time, we jointly minimize the regression loss (cosine proximity) and the sign classification loss (cross-entropy). For the inference, we do the final prediction by merging the absolute values and sign labels together:
\begin{equation}
  \begin{cases}
    cos\phi = sign(cos\phi) \cdot |cos\phi| \\
    sin\phi = sign(sin\phi) \cdot |sin\phi|
  \end{cases}
\end{equation}
Beyond Euler angles, other 2D rotations can be learned in the same fashion.


\partitle{$S^2$ case: Surface normal estimation.}
%
%
A surface normal is the direction that is perpendicular to the tangent plane of the point on the surface of objects in a 3D scene, see Fig. \ref{fig:use_case}.(b).
It can be represented by a unit 3D vector $\mathbf{v}=[N_x,N_y,N_z]$ for which $N_x^2+N_y^2+N_z^2=1$.
Thus, a surface normal lies on the surface of a unit 3D ball, \ie an $S^2$ sphere. Surface normal estimation from RGB images makes pixel-wise predictions of surface normals of the input scene.

It is worth noticing that all surface normals computed by a 2D image should always be pointing outwards from the image plane, that is $N_z<0$, since only these surfaces are visible to the camera. This halves the prediction space to a semi-sphere of $S^2$. Again, when designing the spherical regressor for surface normals, we have a regression branch to learn the absolute normal values $[|N_x|,|N_y|,|N_z|]$ and a classification branch for learning all combinations of signs for $N_x$ and $N_y$.
The total number of possible sign classes is 4, similar to Euler angle estimation.
The training and inference is similar to Euler angles as well. Other $S^2$ problems include learning the direction of motion in 2D/3D flow fields, geographical locations on the Earth sphere and so on.

\partitle{$S^3$ case: 3D rotation estimation.}
Rotational transformations are relevant in many computer vision tasks, for example,
orientation estimation, generalized viewpoint and pose estimation beyond Euler angles or camera relocation.
Rotational transformations can be expressed as orthogonal matrices of size $n$ with determinant +1 (rotation matrices). We can think of the set of all possible rotation matrices to form a group that acts as an operator on vectors. This group is better known as the \emph{special orthogonal Lie group} $\mathcal{SO}(n)$ ~\cite{gurarie1993symmetries}. Specifically, the $\mathcal{SO}(2)$ represents the set of all 2D rotation transformations, whereas $\mathcal{SO}(3)$ represents the set of all possible 3D rotations.

We have already shown that 2D rotations can be mapped to a regression on an $S^1$ sphere, thus the set $\mathcal{SO}(2)$ of all 2D rotations is topologically equivalent to the $S^1$ sphere.
Interestingly, the topology of 3D rotations is not as straightforward~\cite{gurarie1993symmetries}, namely there is no $n$-sphere that is equivalent to $\mathcal{SO}(3)$. Instead, as shown in Fig. \ref{fig:use_case}.(c) a 3D rotation $\mathcal{SO}(3)$ can be thought of as first choosing a rotation axis $\textbf{v}$ and then rotating by an angle $\theta$.
This approach leads to the well known $S^3$ representation of quaternions~\cite{hamilton1844ii}, which is the closest equivalent to the 3D rotation~\cite{shoemake1985animating}.

A unit quaternion is equal to $q = a+b \mathbf{i}+c \mathbf{j}+d \mathbf{k}$, where $a^2+b^2+c^2+d^2=1$.
As $q$ and $-q$ give the same rotation, we restrict ourselves to $a>0$, which again halves the output space.
We, therefore, need to predict the signs of only 3 imaginary components $\{b,c,d\}$ to a total of 8 ($2^3$) classes. The design of the prediction heads and the loss functions are similar to the case of surface normal prediction on $S^2$, only now having 8 sign classes. Given the axis-angle representation ($\theta, \textbf{v}$) of $\mathcal{SO}(3)$, we can, therefore, rewrite a quaternion into $q=(cos\frac{\theta}{2}, sin\frac{\theta}{2}\textbf{v})$. Constraining $a>0$ is equivalent to restricting the rotation angle $\theta \in [0,\pi]$. Furthermore, predicting the 8 sign categories is equivalent to predicting to which of the 8 quadrants of the 3D rotation space the $\textbf{v}$ belongs.


\section{Related work}          \label{sec:related}
\begin{figure}[t!]
\centering
\includegraphics[width=1.0\linewidth]{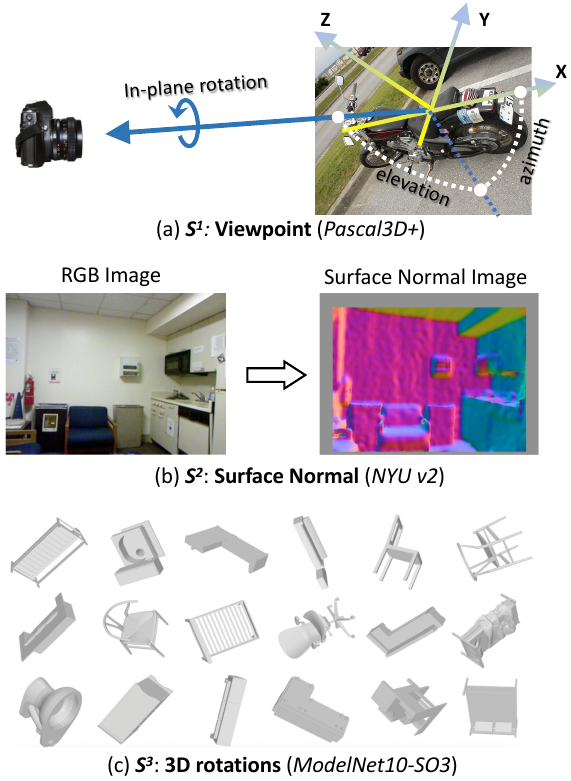}  
   \caption{\textbf{We assess spherical regression on 3 computer vision tasks.}
   (a) $S^1$: Viewpoint estimation on Pascal3D+~\cite{xiang2014beyond}, which needs to predict 3 Euler angles: azimuth, elevation and in-plane rotation.
   (b) $S^2$: Surface normal estimation on NYU v2~\cite{silberman2012indoor}, where pixel-wised dense surface normal prediction is required.
   (c) $S^3$: 3D rotation on our newly proposed ModelNet10-SO3, where given one rendered view of a CAD model, we predict the underlying 3D rotation that aligns it back to standard pose.
   }
\label{fig:3_tasks}
\end{figure}

\partitle{Viewpoint Estimation.}
In general, viewpoint estimation focuses on recovering the 3 Euler angles, namely, azimuth, elevation and in-plane rotation (see Fig. \ref{fig:3_tasks}-(a)). Tulsiani and Malik \cite{tulsiani2015viewpoints} discretize continuous Euler angles into multiple bins and convert viewpoint estimation into a classification problem. Su \etal \cite{su2015render} propose a finer-grained discretization that divides the Euler angles into 360 bins. However, training for all possible outputs requires an enormous amount of examples that can only be addressed by synthetic renderings.

Albeit more natural, regression-based viewpoint estimation is less popular. Because of the periodical nature of angles, most approaches do not regress directly on the linear space of angles, $a, e, t \in [-\pi, \pi]$. The reason is that ignoring the angle periodicity leads to bad modeling, as the $1^\circ$ and $359^\circ$ angles are assumed to be the furthest apart. Instead, trigonometric representations are preferred, with
~\cite{penedones2012improving, beyer2015biternion, prokudin2018deep} proposing to represent angles by $[cos\phi$, $sin\phi]$. They then learn a regression function $h: x \mapsto [cos\phi, sin\phi]$, without, however, enforcing the vectors to lie on $S^1$.
In comparison to viewpoint classification, regression gives continuous and fine-grained angles.
In practice, however, training regression for viewpoint estimation is not as easy.
Complex loss functions are typically crafted, \eg, smooth $L_1$ loss ~\cite{massa2016crafting},
without reaching the accuracy levels of classification-based alternatives.

In this paper, we continue the line of work on regression based viewpoint estimation.
Built upon the $S^1$ representations $[cos\phi$, $sin\phi]$ of Euler angles~\cite{penedones2012improving, beyer2015biternion, prokudin2018deep}, we assess our spherical regression for viewpoint prediction.

\partitle{Surface Normal Estimation.}
Surface normal estimation is typically viewed as a 2.5D representation problem, one that carries information for the geometry of the scene, including layout, shape and even depth. The surface normal is a 3-dim vector that points outside the tangent plane of the surface. In the surface normal estimation task, given an image of a scene, a pixel-wise prediction of the surface normal is required ~\cite{bansal2016marr,eigen2015predicting,zhang2017physically,fouhey2013data,ladicky2014discriminatively,silberman2012indoor,qi2018geonet,wang2014multi} (see Fig. \ref{fig:3_tasks}-(b)).

Fouhey \etal ~\cite{fouhey2013data} infer the surface normal by discovering discriminative and geometrically 3D primitives from 2D images.
Building on contextual and segment-based cues, Ladicky \etal ~\cite{ladicky2014discriminatively} build their surface normal regressor from local image features. They both use hand crafted features.
Eigen and Fergus ~\cite{eigen2015predicting} propose a multi-scale CNN architecture adapted to predicting depth, surface normals and semantic labels. While the network outputs are $\ell_2$ normalized, the gradients are not constrained.
%
Bansal \etal ~\cite{bansal2016marr} introduce a skip-network model optimized by the standard sum of squared errors regression loss,
without enforcing any structure to the output.
%
Zhang \etal ~\cite{zhang2017physically} propose to predict normals with deconvolution layers
and rely on large scale synthetic data for training. Similar to~\cite{eigen2015predicting}, they also enforce an $\ell_2$ norm on the output but have unconstrained gradients.
%
Recently, Qi \etal ~\cite{qi2018geonet} proposed  two-stream CNNs that jointly predict depth and surface normals from a single image and also rely on the sum of squared errors loss for training.

In our work we propose a spherical exponential mapping for performing spherical regression.
This new mapping can be directly applied to any of the surface normal estimation methods that
rely on a regression loss on $n$-spheres and improve their accuracy, as we show in the experiments.

\partitle{3D Rotation Estimation.}
%
%
3D Rotations are a component of several tasks in computer vision and robotics, including viewpoint and pose estimation or camera relocation.
The rotation matrix for 3D rotation is a $3\times3$ orthogonal matrix (determinant$=1$).
Direct regression on the rotation matrix via neural networks is difficult, as the output lies in the $\mathbb{R}^9$ ($3\times3$) space. Moreover, regressing a rotation matrix directly cannot guarantee its orthogonality.
Recently, Falorsi \etal ~\cite{falorsi2018explorations}
take a first step toward regressing 3D rotation matrices. Instead of predicting the 9 elements of rotation matrix directly, they pose the 3D rotation as an $S^2\times S^2$ representation problem reducing the number of elements to regress on to a total of 6.


Viewpoint ~\cite{tulsiani2015viewpoints,su2015render,massa2016crafting,divon2018viewpoint,beyer2015biternion,massa2016crafting,3d_Bbox} and pose ~\cite{osadchy2007synergistic,penedones2012improving} consider the relative 3D rotation between object and camera. With 3 consecutive rotation angles, see Fig. \ref{fig:3_tasks} (a), \textit{Euler Angles} can uniquely recover the rotation matrix. As such a decomposition is easy to be interpreted and able to cover most of the viewpoint distribution, it has been widely adopted.
However, this approach leads to the gimbal lock problem~\cite{hoag1963apollo}, where the degrees of freedom for the rotations are reduced.

Mahendran \etal~\cite{mahendran20173d} studied an axis-angle representation for viewpoint estimation by first choosing a rotation axis and then rotating along it by an angle $\theta$. To constrain the angle $\theta \in [0,\pi)$ and the axis $v_i \in [-1,1]$, they propose a $\pi \cdot tanh$ non-linearity. Also, instead of a standard regression loss, \eg cosine proximity or sum of squared errors loss, they propose a geodesic loss which directly optimizes the 3D rotations in $\mathcal{SO}(3)$.
Do \etal ~\cite{do2018realtime} consider the Lie-algebra $\mathcal{SO}(3)$ representation to learn the 3D rotation of the 6 DoF pose of an object.
It is represented as $[x,y,z] \in R^3$, and can be mapped to a rotation matrix via the Rodrigues rotation formula~\cite{Brockett}. They conclude that an $\ell_1$ regression loss yields better results.

Last, both Kendall \etal ~\cite{kendall2017geometric} and Mahendran \etal ~\cite{mahendran20173d} consider quaternion for camera re-localization and viewpoint estimation.
As quaternions allow for easy interpolation and computations on the $S^3$ sphere, they are also widely used in graphics ~\cite{shoemake1985animating,dam1998quaternions} and robotics ~\cite{mccarthy1990introduction}. Although Do \etal~\cite{do2018realtime} argue that quaternion is over-parameterized, we see this as an advantage that gives us more freedom to learn rotations directly on the $n$-sphere.

Despite the elegance and completeness of the aforementioned works, modelling 3D rotations is hard and  methods specialized for the task at hand, instead, typically reach better accuracies. Unlike most of the aforementioned works, we learn to regress on the Euclidean space directly. Furthermore, we present a framework for regressing on $n$-spheres with constrained gradients, leading to more stable training and good accuracy, as we show experimentally.

\section{Experiments}           \label{sec:experiments}

\subsection{$S^1$: Viewpoint estimation with Euler angles}
\label{ssec:vps_esti}


\partitle{Setup.}
First, we evaluate spherical regression on $S^1$ viewpoint estimation on Pascal3D+~\cite{xiang2014beyond}. Pascal3D+ contains 12 rigid object categories with bounding boxes and noisy rotation matrix annotations, obtained after manually aligning 3D models to the 2D object in the image.
We follow~\cite{tulsiani2015viewpoints, su2015render, prokudin2018deep, mahendran2018mixed, 3d_Bbox} and estimate the 3 Euler angles, namely the azimuth, elevation and in-plane rotation, given the ground truth object location. A viewpoint prediction is correct when the geodesic distance $\Delta(R_{gt}, R_{pr})=\frac{||logR_{gt}^T R_{pr}||_\mathcal{F}}{\sqrt{2}}$  between the predicted rotation matrix $R_{pr}$ (constructed from the predicted Euler angles) and the ground truth rotation matrix $R_{gt}$ is smaller than a threshold $\theta$~\cite{tulsiani2015viewpoints}. The evaluation metric is the accuracy $Acc@\pi/6$ given threshold $\theta=\pi/6$. We use ResNet101 as our backbone architecture, with a wider penultimate fully connected layer in the prediction head that is shared by the regression branch and classification branch (see supplementary material for details).
As many of the annotations are concentrated around the $x$-axis, we found that rotating all annotations by 45$^\circ$ during training (and rotating back at test time) leads to more balanced distribution of annotations and better learning. 
For training data, we also use the synthetic data provided by~\cite{su2015render}, without additional data augmentations like in~\cite{mahendran20173d,mahendran2018mixed}.

\begin{table}[t!]
\centering
\caption{\textbf{$S^1$: Viewpoint estimation with Euler angles.} Comparison with state-of-the-art on Pascal3D+. Adding our $S_{exp}^1$ spherical regression on top of the backbone network of~\cite{penedones2012improving} leads to best accuracy. We report a class-wise comparison in supplementary.
%
%
         }
\label{table:pascal3d-sota}
\begin{tabular}{lcc}
\toprule
 &   \textbf{MedErr\scriptsize$\downarrow$} & \textbf{Acc@$\frac{\pi}{6}$\scriptsize$\uparrow$}    \\
\midrule
Mahendran \etal    ~\cite{mahendran20173d}            &          16.6        &          N/A             \\
Tulsiani and Malik ~\cite{tulsiani2015viewpoints}     &          13.6        &         80.8             \\
Mousavian \etal    ~\cite{3d_Bbox}                    &          11.1        &         81.0             \\
Su \etal           ~\cite{su2015render}               &          11.7        &         82.0             \\
Penedones \etal    ~\cite{penedones2012improving}\dag
                                                      &          11.6        &         83.6             \\
Prokudin \etal     ~\cite{prokudin2018deep}           &          12.2        &         83.8             \\
Grabner  \etal     ~\cite{grabner20183d}              &          10.9        &         83.9             \\
Mahendran \etal    ~\cite{mahendran2018mixed}         &          10.1        &         85.9             \\
\midrule
\textit{This paper:} \cite{penedones2012improving}\dag + $S_{exp}^1$
                                                   &       \B{ 9.2}       &      \B{88.2}                  \\
\bottomrule
\dag \,\scriptsize{Based on our implementation.}  \\
\end{tabular}
\end{table}

\bigbreak\partitle{Results.}
We report comparisons with the state-of-the-art in Table~\ref{table:pascal3d-sota}.
Note that our spherical exponential mapping can be easily used by any of the regression-based methods with $S^1$ representation $[cos\phi,sin\phi]$ ~\cite{penedones2012improving, beyer2015biternion}.
In this experiment we combine it with Penedones \etal ~\cite{penedones2012improving}, who tried to directly regress 2D representation [$cos \phi$, $sin \phi$] of angels, obtaining a significant improvement in accuracy over other regression and classification baselines.
That said, during experiments we observed that classification-based methods are more amenable to large data sets, most probably because of their increased number of parameters.
As expected, the continuous outputs by the spherical regression are better suited for finer and finer evaluations , that is $Acc@\pi/12$ and $Acc@\pi/24$ (supplementary material).
We conclude that spherical regression is successful for viewpoint estimation with Euler angles.

\subsection{$S^2$: Surface normal estimation}

\partitle{Setup.} Next, we evaluate spherical regression for $S^2$ surface normal estimation on the NYU Depth v2~\cite{silberman2012indoor}. The NYU Depth v2 dataset contains 1,449 video frames of indoor scenes associated with Microsoft Kinect depth data. We use the ground truth surface normals provided by~\cite{silberman2012indoor}. We consider all valid pixels across the whole test set during evaluation~\cite{zhang2017physically}. The evaluation metrics are the (\textit{Mean} and \textit{Median}), as well as the accuracy based metric, namely the percentage of correct predictions at given threshold $\mathit{11.24^\circ}$, $\mathit{22.5^\circ}$ and $\mathit{30^\circ}$).
We implement our $S_{exp}^2$ spherical regression based on the network proposed by Zhang \etal ~\cite{zhang2017physically}, which is built on top of VGG-16 convolutional layers, and a symmetric stack of deconvolution layers with skip connections for decoding.
%
%
As in viewpoint estimation, we also rotate the ground truth around the $z$-axis by 45$^\circ$ to yield better results. We follow the same training setup as~\cite{zhang2017physically}, that is we first pre-train on the selected 568K synthetic data provided by~\cite{zhang2017physically} for 8 epochs, and fine-tune on NYU v2 for 60 epochs.

\begin{table}[t!]
\centering
\caption{\textbf{$S^2$: Surface normal estimation} Comparison with state-of-the-art on NYU v2. Adding our $S_{exp}^2$ spherical regression on top of the backbone network of Zhang \etal \cite{zhang2017physically} leads to best accuracy.}
\label{table:surface-normals}
\begin{adjustbox}{max width=\linewidth}
\setlength\tabcolsep{1.5pt} 
    \begin{tabular}{lrrrrr}
    \toprule
        &  \textbf{Mean\scriptsize$\downarrow$} & \textbf{Median\scriptsize$\downarrow$} & \textbf{$11.25^\circ$\scriptsize$\uparrow$} & \textbf{$22.5^\circ$\scriptsize$\uparrow$} & \textbf{$30.0^\circ$\scriptsize$\uparrow$}  \\
    \midrule
     Fouhey \etal ~\cite{fouhey2013data} \S                 & 37.7 & 34.1 & 14.0 & 32.7 & 44.1 \\
     Ladicky \etal ~\cite{ladicky2014discriminatively} \S   & 35.5 & 25.5 & 24.0 & 45.6 & 55.9 \\
     Wang \etal ~\cite{wang2014multi} \S                    & 28.8 & 17.9 & 35.2 & 57.1 & 65.5 \\
     Eigen and Fergus~\cite{eigen2015predicting}            & 22.3 & 15.3 & 38.6 & 64.0 & 73.9 \\
     Zhang \etal~\cite{zhang2017physically}                 & 21.7 & 14.8 & 39.4 & 66.3 & 76.1 \\
     \midrule
     \textit{This paper}: \cite{zhang2017physically} + $S_{exp}^2$
                                            & {\B{19.7}}  &  {\B{12.5}}  &  {\B{45.8}}  &  {\B{72.1}}  &  {\B{80.6}}  \\
    \bottomrule
      \S \,\scriptsize{Copied from \cite{eigen2015predicting}.}  \\
    \end{tabular}
\end{adjustbox}
\end{table}

\begin{table}[t!]
\centering
\caption{\textbf{$S^3$: 3D Rotation estimation with quaternions.}  Comparison on newly established ModelNet10-SO3. Adding our $S_{exp}^3$ spherical regression on top of an AlexNet or VGG16 backbone network leads to best accuracy.}
\label{table:modelnet10_SO3}
\resizebox{1.0\columnwidth}{!}{%
\setlength\tabcolsep{0.5pt} 
    \begin{tabular}{lrrrr}
    \toprule
    &  \textbf{\small MedErr\scriptsize$\downarrow$} & \textbf{\small Acc@$\frac{\pi}{6}$\scriptsize$\uparrow$} & \textbf{\small Acc@$\frac{\pi}{12}$\scriptsize$\uparrow$} & \textbf{\small Acc@$\frac{\pi}{24}$\scriptsize$\uparrow$}  \\
    \midrule
    AlexNet (Direct+smooth-L1) &   46.1   &  32.5   &  11.2   &    2.5   \\
    AlexNet + $S_{flat}$       &   33.3   &  53.5   &  34.1   &   13.9   \\ 
    AlexNet + $S_{exp}^3$      &\B{25.3}  &\B{65.4} &\B{48.5}&\B{24.4}   \\
    \hline
    VGG16 (Direct+smooth-L1)   &   36.8   &  46.7   &  29.4   &   13.4   \\ 
    VGG16   + $S_{flat}$       &   25.9   &  63.5   &  48.7   &   29.5   \\ 
    VGG16   + $S_{exp}^3$      &\B{20.3} &\B{70.9} &\B{58.9}  &\B{38.4}  \\ 
    \bottomrule
    \end{tabular}
}
\end{table}


\partitle{Results.}
We report results in Table~\ref{table:surface-normals}. Replacing regular regression in~\cite{zhang2017physically} with spherical regression on $S^2$ improves the estimation of the surface normals considerably.
We found the improvement is attentuated by the fact that for surface normal estimation we perform one regression per pixel location.
As each one of these regressions could return unstable gradients, bounding the total sum of losses with spherical regression is beneficial. Especially for the finer regression thresholds of $11.25^\circ, 22.5^\circ$. We conclude that spherical regression is successful also for surface normal estimation.

\subsection{$S^3$: 3D Rotation estimation with quaternions}

\partitle{Setup.}
Last, we evaluate $S_{exp}^3$ spherical regression on 3D rotation estimation on $S^3$ with quaternions.
For this evaluation we introduce a new dataset, \textit{ModelNet10-SO3}, composed of images of 3D synthetic renderings. ModelNet10-SO3 is based on \textit{ModelNet10}~\cite{wu20153d}, which contains 4,899 instances from 10 categories of 3D CAD models. In ModelNet10 the purpose is the classification of 3D shapes to one of the permissible CAD object categories.
With ModelNet10-SO3 we have a different purpose, we want to evaluate 3D shape alignment by predicting its 3D rotation matrix \wrt the reference position from single image. We construct ModelNet10-SO3 by uniformly sampling per CAD model 100 3D rotations on $\mathcal{SO}(3)$ for the training set and 4 3D rotations for the test set. We render each view with white background, thus the foreground shows the rotated rendering only. We show some examples in Fig.~\ref{fig:3_tasks}-(c).

Relying on Euler angles for ModelNet10-SO3 is not advised because of the Gimbal lock problem~\cite{hoag1963apollo}.
Instead, alignment is possible only by predicting the quaternion representation of the 3D rotation matrix.
For this task, we test the following 3 regression strategies:

\begin{itemize} 
\itemsep1pt  
\item[(I)]    Direct regression with smooth-L1 loss. It may cause the output to no longer follow unit $\ell_2$ norm.
\item[(II)]   Regression with $\ell_2$ normalization $\mathcal{S}_{flat}$.
\item[(III)]  Regression with $\mathcal{S}_{exp}$ (this paper).
\end{itemize}

We report results based on AlexNet and VGG16 as our CNN backbones, with a class-specific prediction head.
We borrow the evaluation metric from viewpoint estimation, namely \textit{MedErr} and \textit{$Acc@\{\pi/6, \pi/12, \pi/24\}$} so that we also examine finer-grained predictions.

\partitle{Results.}
We report results in Table \ref{table:modelnet10_SO3}. First, both $S_{flat}$ and $S_{exp}^3$ regression on quaternions improve over direct regression baselines.
This shows the importance of constraining the output space to be on sphere when regress spherical target. Second, putting $l_2$ normalization constraint on output space, $S_{exp}^3$ improves over $S_{flat}$ with both AlexNet and VGG16.
For AlexNet we obtain about $8-12\%$ improvement across all metrics. VGG16 is higher overall, but the improvement over the baseline is less.
This shows that with the VGG16 we are potentially getting closer to the maximum possible accuracy attainable for this hard task.
That can be explained by the fact that the shapes have no texture. Thus, a regular VGG16 is close to what can be encoded by a good RGB-based model.
Note that estimating the 3D rotation with a discretization and classification approach~\cite{tulsiani2015viewpoints, su2015render, massa2016crafting, divon2018viewpoint} would be impossible because of the vastness of the output space on $\mathcal{SO}(3)$ manifold.

\begin{figure}[t!]
\begin{center}
   \includegraphics[width=1.0\linewidth]{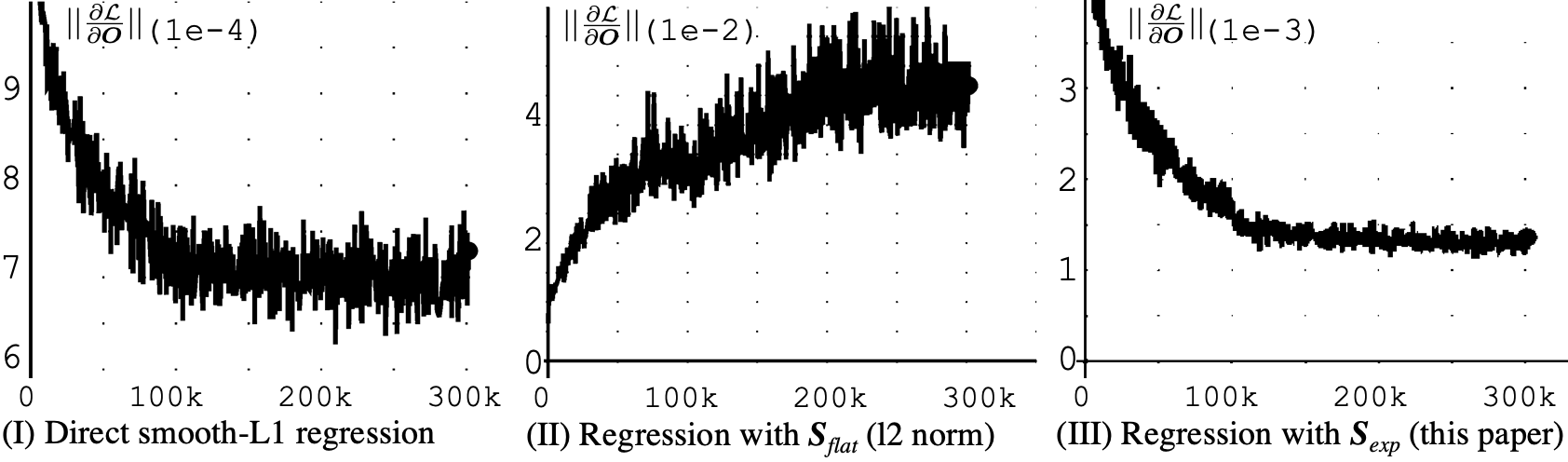}
\end{center}
    \vspace{-1em}
   \caption{\textbf{Variance of the average gradient norm $||\frac{\partial \mathcal L}{\partial \vt O}||$.} Spherical exponentiation $\mathcal{S}_{exp}$ yields lower variance on mini-batch over entire train progress.}
\label{fig:avg_gradient}
\end{figure}

Further, we investigate the variance of the gradients $\frac{\partial \mathcal L}{\partial \vt O}$ by recording its average $\ell_2$ norm during training progress. The results are shown in Fig. \ref{fig:avg_gradient}. We observe the gradient norm of the spherical exponential mapping has much lower variance. Spherical exponentiation achieves this behavior naturally without interventions, unlike other tricks (\eg gradient clipping, gradient reparameterization) which fix the symptom (gradient instability/vanishing/exploding) but not the root cause (unconstrained input signals). 
We conclude that spherical regression is successful also for the application of 3D rotation estimation.

\section{Conclusion}            \label{sec:conclusions}


Spherical regression is a general framework which can be applied to any continuous output problem that lives in $n$-spheres. It obtains regression gradients that are constrained and well-behaving for several computer vision challenges. In this work we have investigated three such applications, specifically viewpoint estimation, surface normal estimation and 3D rotation estimation.
Generally, we observe that spherical regression improves considerably the regression accuracy in all tasks and different datasets.
We conclude that spherical regression is a good alternative for tasks where continuous output prediction are needed.


{\small
\bibliographystyle{ieee_fullname}
\bibliography{main}
}

\end{document}


\title{Supplementary Material\\
       Spherical Regression:\\ Learning Viewpoints, Surface Normals and 3D Rotations on \emph{n}-Spheres}

\author{
    Shuai Liao \\
    \and
    Efstratios Gavves\\
    \and
    Cees G. M. Snoek\\
    \and
    \and
    QUVA Lab, University of Amsterdam
}

\maketitle
\thispagestyle{empty}


\section{$S^1$: Viewpoint estimation with Euler angles}
\begin{figure*}[hbt!]
\begin{center}
\includegraphics[width=0.7\linewidth]{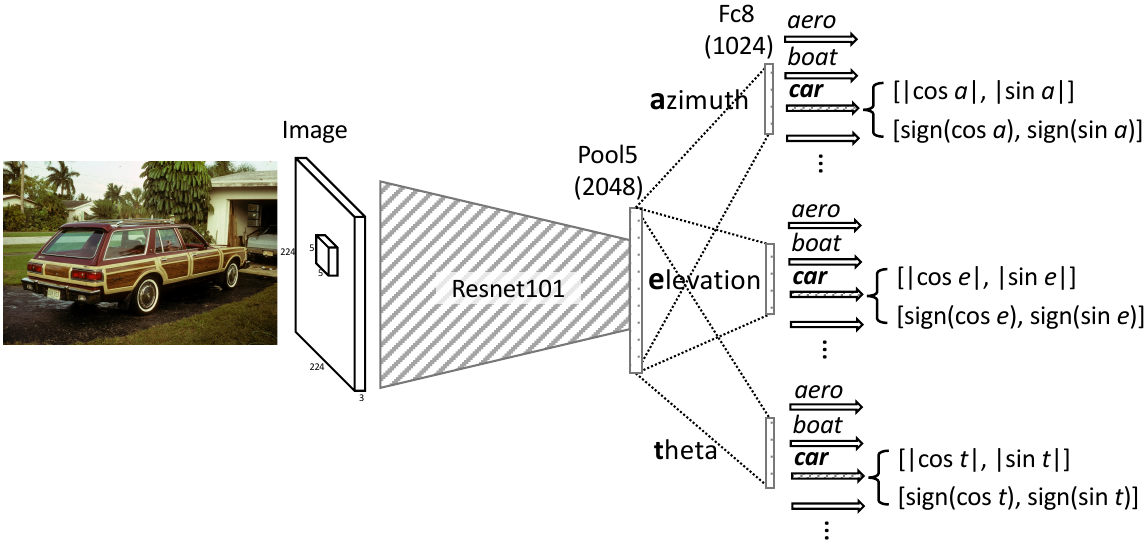}
\end{center}
   \caption{\textbf{Network architecture for viewpoint estimation by Euler angles on Pascal3D+}.}
\label{fig:system}
\end{figure*}

%
We show the viewpoint estimation network architecture used in this paper in Fig. \ref{fig:system}.
Given ResNet101 as backbone to provide a shared Pool5 feature (with 2048 output unit),
we have 3 branches to estimate azimuth, elevation and in-plane rotation (theta) angels.
Each branch begins with a fully-connected layer (Fc8), with 1024 output units, and makes a prediction for the 12 categories in Pascal3D+.
Our prediction head is composed of two components: 1) absolute value prediction and 2) sign prediction.

We show the fine-grained evaluation in Table. \ref{table:vps_finer}.
In comparison with Penedones \etal ~\cite{penedones2012improving},
spherical regression improves the performance in all evaluation metrics, namely Acc@$\{\frac{\pi}{6},\frac{\pi}{12},\frac{\pi}{24}\}$.

We report the class-wise performance comparison in Table. \ref{tbl:pascal3d_class_wise}.
Prokudin \etal  ~\cite{prokudin2018deep} wins the most categories under MedError metric (5 out of 12).
However, they made a larger mistake on difficult categories like boat, where the visual appearance has larger variance.
For Acc@$\frac{\pi}{6}$ metric, our method wins the most (6 out of 12 categories).
In comparison with Penedones \etal ~\cite{penedones2012improving},
adding spherical regression module consistently helps increase the accuracy across almost all categories.

\begin{table*}[h]
\centering
\caption{ \textbf{Viewpoint estimation with fine-grained evaluation on Pascal3D+}.
          We report results of \textbf{Acc@$\{\frac{\pi}{6},\frac{\pi}{12},\frac{\pi}{24}\}\uparrow$}.
          Results generated by spherical regression module ($S_{exp}^3$) have a better alignment to the ground truth models.
         }
\label{table:vps_finer}
\begin{adjustbox}{max width=0.6\linewidth}
\begin{tabular}{lcccc}
\toprule
 &   \textbf{MedErr$\downarrow$} & \textbf{Acc@$\frac{\pi}{6}\uparrow$}  & \textbf{Acc@$\frac{\pi}{12}\uparrow$}  & \textbf{Acc@$\frac{\pi}{24}\uparrow$}  \\
\midrule
Penedones \etal ~\cite{penedones2012improving}\dag
                                                   &   11.6   &   83.6     &   66.3    &   35.9      \\
\midrule
\textit{This paper:} \cite{penedones2012improving}\dag + $S_{exp}^1$
                                                   &\B{ 9.2}  &\B{88.2}    &\B{74.1}   & \B{46.0}    \\
\bottomrule
\dag \,\footnotesize{Based on our implementation.}  \\
\end{tabular}
\end{adjustbox}
\end{table*}

\noindent
\begingroup
\setlength{\tabcolsep}{3pt} 
\begin{table*}[h]
\centering
\caption{\textbf{Category-wise evaluation of viewpoint estimation on Pascal3D+}.
        }
\begin{adjustbox}{max width=0.83\linewidth}
\begin{tabular}{llrrrrrrrrrrrrr}
\toprule 
  & Method&   aero &   bike &   boat & bottle &    bus &    car &  chair &  table &  mbike &   sofa &  train &     tv &   mean   \\
\midrule 
{\multirow{6}{*}{\rotatebox[origin=c]{90}{MedError}}}
  & Mahendran \etal ~\cite{mahendran20173d}                &   14.5 &   22.6 &   35.8 &    9.3 &    4.3 &    8.1 &   19.1 &   30.6 &   18.8 &   13.2 &    7.3 &   16.0 &   16.6   \\
  & Tulsiani  \etal ~\cite{tulsiani2015viewpoints}         &   13.8 &   17.7 &   21.3 &   12.9 &    5.8 &    9.1 &   14.8 &   15.2 &   14.7 &   13.7 &    8.7 &   15.4 &   13.6   \\
  & Mousavian \etal ~\cite{3d_Bbox}                        &   13.6 &   12.5 &   22.8 &    8.3 &    3.1 &    5.8 &   11.9 &   12.5 &   12.3 &   12.8 &    6.3 &   11.9 &   11.1   \\
  & Su \etal        ~\cite{su2015render}                   &   15.4 &   14.8 &   25.6 &    9.3 &    3.6 &    6.0 &    9.7 &   10.8 &   16.7 &    9.5 &    6.1 &   12.6 &   11.7   \\
  & Penedones \etal ~\cite{penedones2012improving}\dag     &   12.3 &\B{11.5}&  31.3  &    6.9 &    4.4 &    7.1 &   12.2 &   13.9 &   13.1 & \B{7.7}&    7.0 &   12.1 &   11.6   \\
  & Prokudin  \etal ~\cite{prokudin2018deep}               &    9.7 &   15.5 &   45.6 &\B{ 5.4}& \B{2.9}&\B{ 4.5}&   13.1 &   12.6 &\B{11.8}&    9.1 &\B{ 4.3}&   12.0 &   12.2   \\
  & Grabner   \etal ~\cite{grabner20183d}                  &   10.0 &   15.6 &\B{19.1}&    8.6 &    3.3 &    5.1 &   13.7 &   11.8 &   12.2 &   13.5 &    6.7 &   11.0 &   10.9   \\
  & Mahendran \etal ~\cite{mahendran2018mixed}             &\B{8.5} &   14.8 &   20.5 &    7.0 &    3.1 &    5.1 &    9.3 &   11.3 &   14.2 &   10.2 &    5.6 &   11.7 &   10.1   \\
  & \textit{This paper:} ~\cite{penedones2012improving}\dag + $S_{exp}^1$
                                            &    9.2 &   11.6 &   20.6 &    7.3 &    3.4 &    4.8 &\B{ 8.2}&\B{ 8.5}&   12.1 &    8.7 &    6.1 &\B{10.1}&\B{ 9.2}  \\
\midrule 
\midrule 
{\multirow{6}{*}{\rotatebox[origin=c]{90}{$Acc@\pi/6$}}}
 & Mahendran \etal ~\cite{mahendran20173d}                &    N/A &    N/A &    N/A &    N/A &    N/A &     N/A &    N/A &    N/A &    N/A &    N/A &    N/A &    N/A &      N/A \\
 & Tulsiani  \etal ~\cite{tulsiani2015viewpoints}         &   0.81 &   0.77 &   0.59 &   0.93 &\B{0.98}&    0.89 &   0.80 &   0.62 &   0.88 &   0.82 &   0.80 &   0.80 &   0.808 \\ 
 & Mousavian \etal ~\cite{3d_Bbox}                        &   0.78 &   0.83 &   0.57 &   0.93 &   0.94 &    0.90 &   0.80 &   0.68 &   0.86 &   0.82 &   0.82 &   0.85 &   0.810 \\ 
 & Su \etal        ~\cite{su2015render}                   &   0.74 &   0.83 &   0.52 &   0.91 &   0.91 &    0.88 &   0.86 &   0.73 &   0.78 &   0.90 &\B{0.86}&   0.92 &   0.820 \\ 
 & Penedones \etal ~\cite{penedones2012improving}\dag     &   0.80 &   0.85 &   0.48 &\B{0.96}&   0.94 &    0.91 &   0.84 &   0.70 &   0.86 &   0.95 &   0.84 &   0.91 &   0.836 \\ 
 & Prokudin  \etal ~\cite{prokudin2018deep}               &\B{0.89}&   0.83 &   0.46 &\B{0.96}&   0.93 &    0.90 &   0.80 &\B{0.76}&   0.90 &   0.90 &   0.82 &   0.91 &   0.838 \\ 
 & Grabner   \etal ~\cite{grabner20183d}                  &   0.83 &   0.82 &\B{0.64}&   0.95 &   0.97 &    0.94 &   0.80 &   0.71 &   0.88 &   0.87 &   0.80 &   0.86 &   0.839 \\ 
 & Mahendran \etal ~\cite{mahendran2018mixed}             &   0.87 &   0.81 &\B{0.64}&\B{0.96}&   0.97 & \B{0.95}&   0.92 &   0.67 &   0.85 &   0.97 &   0.82 &   0.88 &   0.859 \\ 
 & \textit{This paper:} ~\cite{penedones2012improving}\dag + $S_{exp}^1$  &   0.88 &\B{0.88}&   0.61 &\B{0.96}&   0.97 &    0.93 &\B{0.93}&   0.74 &\B{0.93}&\B{0.98}&   0.84 &\B{0.95}&\B{0.882}\\ 
\bottomrule 
& \dag \,\footnotesize{Based on our implementation.}  \\
\end{tabular}
\end{adjustbox}
\label{tbl:pascal3d_class_wise}
\end{table*}
\endgroup

\section{$S^2$: Surface normal estimation}

\begin{figure*}[h]
\begin{center}
\includegraphics[width=0.83\linewidth]{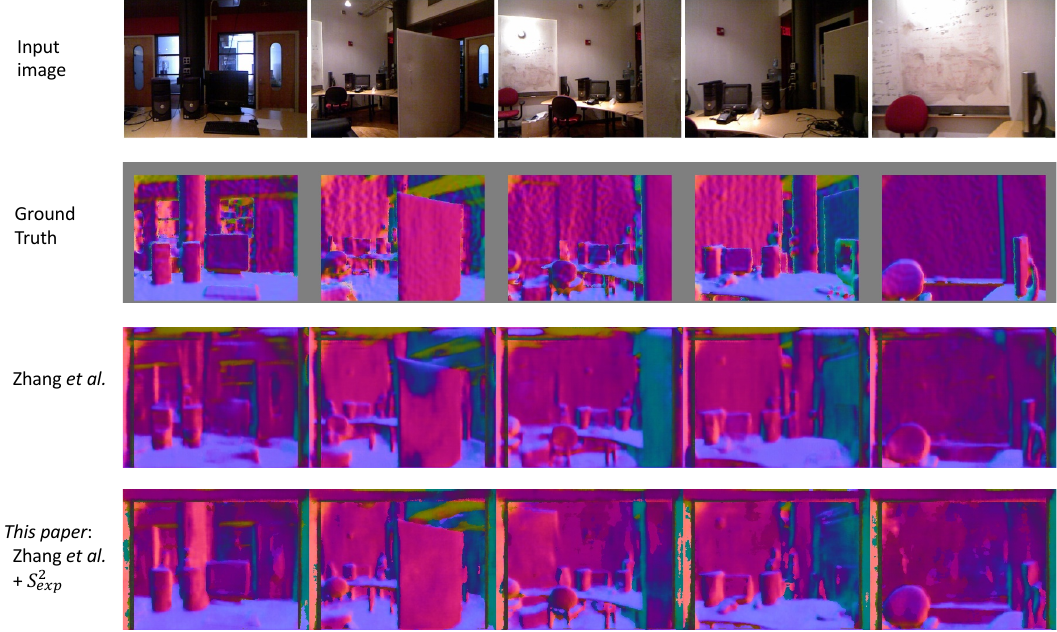}
\end{center}
   \caption{\textbf{Visualization of Surface Normal Estimation on \textit{NYU v2}}.
            Predictions are made by model: ``Zhang \etal ~\cite{zhang2017physically}'' and ``Zhang \etal ~\cite{zhang2017physically} + $S_{exp}^2$''.
            While results from Zhang \etal ~\cite{zhang2017physically} are smoother, our method generates sharp boundaries and thus maintains details.}
\label{fig:vis_surface_normal}
\end{figure*}

We show the visualization of surface normal prediction in Fig. \ref{fig:vis_surface_normal}.
The results from Zhang \etal ~\cite{zhang2017physically} are smoother than our results from spherical regression,
but it makes some mistake with quite large surface area, \eg the wall on the picture at row 3 column 2.
In terms of boundaries, our results tend to be sharper. This is mainly due to the classification branch,
which forces the prediction to choose the main direction in one out of four quadrants.
Overall, our results maintain more details than Zhang \etal ~\cite{zhang2017physically}.

\section{$S^3$: 3D Rotation estimation with quaternions}

We show a class-wise performance comparison based on Acc@$\frac{\pi}{6}$ in Fig. \ref{fig:rot3_class_wise}.
Since we are predicting the 3D rotation just from a single image,
it can be seen that categories with high degree of symmetry have worse performance, \eg bathtub, desk, night-stand and table.
In comparison with the regression of quaternion with flat VGG16, spherical regression consistently helps increase the accuracy.

\begin{figure*}[h]
\begin{center}
\includegraphics[width=0.85\linewidth]{images/Acc@pi12.pdf}  
\end{center}
   \caption{\textbf{Class-wise comparison of 3D rotation estimation on \textit{ModelNet10-SO3}}.
            Categories with high degree of symmetry are observed to have worse performance, \eg bathtub, desk, night-stand and table.
            Spherical regression module ($S_{exp}^3$) consistently helps increase the performance over flat regression of quaternion by VGG16.}
\label{fig:rot3_class_wise}
\end{figure*}

\begin{figure*}[h]
\begin{center}
\includegraphics[width=\linewidth]{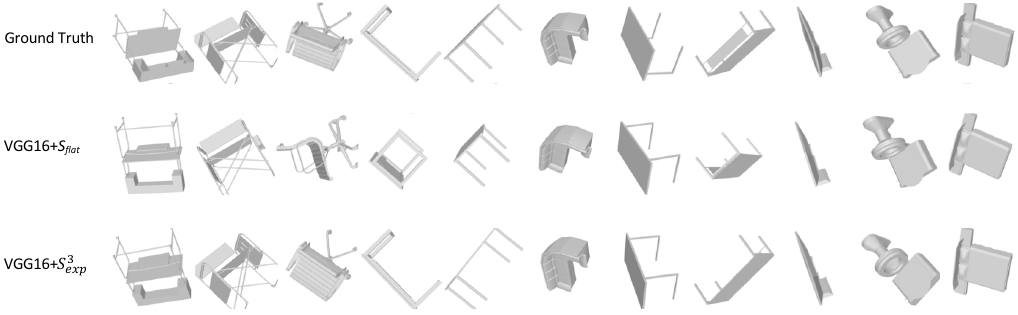}  
\end{center}
   \caption{\textbf{Visualization of 3D rotation estimation on \textit{ModelNet10-SO3}}.}
\label{fig:vis_rot3}
\end{figure*}

We show a visualization of 3D rotation estimation in Fig. \ref{fig:vis_rot3}.
The first row is the ground truth input images.
We render the predicted rotations from VGG16 and VGG16+$S_{exp}^3$ in second and third rows.
We can see our result have a better alignment to the ground truth models.

{}

\section{Derivation of Jacobian for $S_{flat}$ and $S_{exp}$}.

First, we provide detailed derivation of Eq. 7 in the main paper.
Given the $\ell_2$ normalization form: $$p_j =g(o_j; \vt O)= \frac{f(o_j)}{\sqrt{\sum_k f(o_k)^2}}$$ with arbitrary univariate mapping $f(\cdot)$,
we have: 

\begin{align}
  \frac{\partial p_j}{\partial o_i} &= \frac{\frac{d f(o_j)}{d o_i}\cdot A  - f(o_j)\cdot \frac{\partial A}{\partial o_i}               }{A^2} \\
                           &= \frac{\frac{d f(o_j)}{d o_i}\cdot A  - f(o_j)\cdot p_i \cdot \frac{d f(o_i)}{d o_i}}{A^2}  \\
                           &= \frac{1}{A} \left[ \frac{d f(o_j)}{d o_i} - p_i\cdot p_j \cdot \frac{d f(o_i)}{d o_i} \right] \\
                           &= \begin{cases*}
                                 \frac{f'(o_i)}{A} \cdot (1-p_i \cdot p_j),         & \text{when j=i} \\ \frac{f'(o_i)}{A} \cdot (0-p_i \cdot p_j),  & \text{when j$\neq$ i} \\
                              \end{cases*}
\end{align}
where $A=\sqrt{\sum_k f(o_k)^2}$.

%

Thus the Jacobian matrix of $\mathit{g}: \vt{O} \rightarrow \vt{P}$ is as follows

\begin{align}
\mathbf{J}_g &= \fracpar{\vt{P}}{\vt{O}}
  =\left[\fracpar{\vt P}{o_0}, \fracpar{\vt P}{o_1}, \cdots, \fracpar{\vt P}{o_n} \right] \\
  & = \begin{bmatrix}
  \fracpar{p_0}{o_0} &  \fracpar{p_0}{o_1} &\cdots & \fracpar{p_0}{o_n} \\
    \fracpar{p_1}{o_0} &  \fracpar{p_1}{o_1} &\cdots & \fracpar{p_1}{o_n} \\
  \vdots  & \vdots & \ddots  & \vdots  \\
   \fracpar{p_n}{o_0} &  \fracpar{p_n}{o_1} &\cdots & \fracpar{p_n}{o_n} \\
\end{bmatrix} \\
& = \begin{bmatrix}
  1-p_0p_0 &    -p_1p_0 & \cdots & -p_np_0    \\
     -p_0p_1 & 1-p_1p_1 & \cdots & -p_np_1    \\
      \vdots  & \vdots      & \ddots  & \vdots     \\
    -p_0p_n &  -p_1p_n   & \cdots  & 1-p_np_n \\
\end{bmatrix}
\begin{bmatrix}
 \frac{f'(o_0)}{A} &        &              &          \\
         & \frac{f'(o_1)}{A} &              &          \\
         &        & \ddots &           \\
         &        &              & \frac{f'(o_n)}{A}    \\
\end{bmatrix} \\
& = \left( \vt{I} - \begin{bmatrix}
     p_0p_0 &  p_1p_0 & \cdots & p_np_0    \\
     p_0p_1 &  p_1p_1 & \cdots & p_np_1    \\
      \vdots  & \vdots  & \ddots  & \vdots     \\
     p_0p_n &  p_1p_n & \cdots  & p_np_n    \\
\end{bmatrix} \right)
\begin{bmatrix}
 \frac{f'(o_0)}{A} &        &              &          \\
         & \frac{f'(o_1)}{A} &              &          \\
         &        & \ddots &           \\
         &        &              & \frac{f'(o_n)}{A}    \\
\end{bmatrix}\label{eq:general_parfrac}
\end{align}

\subsection{$\mathcal{S}_{flat}$ case}

In this case, we only take flat $\ell_2$ normalization on $\vt O$ to obtain $\vt P$, namely $p_j =g(o_j; \vt O)= \frac{o_j}{\sqrt{\sum_k o_k^2}}$.
This means $f(o_i) = o_i$ and $f'(o_i)=1$. Thus Eq. \ref{eq:general_parfrac} becomes:


\begin{align}
&\mathbf{J}_{\mathcal{S}_{flat}} = \fracpar{\vt{P}}{\vt{O}} \\
& = \left( \vt{I} - \begin{bmatrix}
     p_0p_0 &  p_1p_0 & \cdots & p_np_0    \\
     p_0p_1 &  p_1p_1 & \cdots & p_np_1    \\
      \vdots  & \vdots  & \ddots  & \vdots     \\
     p_0p_n &  p_1p_n & \cdots  & p_np_n    \\
\end{bmatrix} \right)
\begin{bmatrix}
 \frac{1}{A} &        &              &          \\
         & \frac{1}{A} &              &          \\
         &        & \ddots &           \\
         &        &              & \frac{1}{A}    \\
\end{bmatrix} \\
& =[\fracpar{\vt P}{o_0}, \fracpar{\vt P}{o_1}, \cdots, \fracpar{\vt P}{o_n}] \\
& = (\vt{I} - \vt{P}\otimes\vt{P}) \cdot \frac{1}{A}
\end{align}
where $\otimes$ denotes outer product.

\subsection{$\mathcal{S}_{exp}$ case}

In this case, we take spherical normalization on $\vt O$ to obtain $\vt P$, namely $p_j =g(o_j; \vt O)= \frac{e^{o_j}}{\sqrt{\sum_k (e^{o_k})^2}}$.
This means $f(o_i) = e^{o_i}$ and $f'(o_i)=e^{o_i}$. Thus Eq. \ref{eq:general_parfrac} becomes:


\begin{align}
& \mathbf{J}_{\mathcal{S}_{exp}} = \fracpar{\vt{P}}{\vt{O}} \\
& = \left( \vt{I} - \begin{bmatrix}
     p_0p_0 &  p_1p_0 & \cdots & p_np_0    \\
     p_0p_1 &  p_1p_1 & \cdots & p_np_1    \\
      \vdots  & \vdots  & \ddots  & \vdots     \\
     p_0p_n &  p_1p_n & \cdots  & p_np_n    \\
\end{bmatrix} \right)
\begin{bmatrix}
  p_0 &        &              &          \\
         & p_1 &              &          \\
         &        & \ddots &           \\
         &        &              & p_n    \\
\end{bmatrix} \\
&=(\vt{I} - \vt{P}\cdot\vt{P}^T)\cdot diag(\vt{P}) \\
&=(\vt{I} - \vt{P}\otimes\vt{P})\cdot diag(\vt{P})
\end{align}
where $\otimes$ denotes outer product.

{\small
\bibliographystyle{ieee_fullname}
\bibliography{main}
}